\documentclass{article}
\usepackage{spconf,amsmath,epsfig}
\usepackage{amsmath,epsfig}
\usepackage{cite}
\usepackage{color}
\usepackage{amsfonts}
\usepackage[misc]{ifsym} 
\usepackage{underscore}
\usepackage{lipsum}
\let\OLDthebibliography\thebibliography
\renewcommand\thebibliography[1]{
  \OLDthebibliography{#1}
  \setlength{\parskip}{0pt}
  \setlength{\itemsep}{0pt plus 0.3ex}
}

\begin{document}\sloppy

% Example definitions.
% --------------------
\def\x{{\mathbf x}}
\def\L{{\cal L}}

% Title.
% ------
\title{MTNet: Learning modality-aware representation with transformer for RGBT tracking}
%
% Single address.
% ---------------
\name{Ruichao Hou, Boyue Xu, Tongwei Ren$^{(\textrm{\Letter})}$, Gangshan Wu}
\address{State Key Laboratory for Novel Software Technology, Nanjing University, Nanjing, China\\ \{rc_hou, xuby\}@smail.nju.edu.cn, \{rentw, gswu\}@nju.edu.cn }
%Address and e-mail should NOT be added in the submission paper. They should be present only in the camera ready paper. 

\maketitle
\pagestyle{empty}  % no page number for the second and the later pages
\thispagestyle{empty} % no page number for the first page
\newcommand\blfootnote[1]{%
\begingroup
\renewcommand\thefootnote{}\footnote{#1}%
\addtocounter{footnote}{-1}%
\endgroup
}

\blfootnote{$^{(\textrm{\Letter})}$Corresponding Author. This work is supported by the National Science Foundation of China (62072232), the program B for Outstanding Ph.D. candidate of Nanjing University, and the Collaborative Innovation Center of Novel Software Technology and Industrialization.} %无标号，显示

\begin{abstract}
The ability to learn robust multi-modality representation has played a critical role in the development of RGBT tracking. However, the regular fusion paradigm and the invariable tracking template remain restrictive to the feature interaction. In this paper, we propose a modality-aware tracker based on transformer, termed MTNet. Specifically, a modality-aware network is presented to explore modality-specific cues, which contains both channel aggregation and distribution module(CADM) and spatial similarity perception module (SSPM). A transformer fusion network is then applied to capturing global dependencies to reinforce instance representations. To estimate the precise location and tackle the challenges, such as scale variation and deformation, we design a trident prediction head and a dynamic update strategy which jointly maintain a reliable template for facilitating inter-frame communication. Extensive experiments validate that the proposed method achieves satisfactory results compared with the state-of-the-art competitors on three RGBT benchmarks while reaching real-time speed. 
\end{abstract}
\vspace{-0.2cm}
\section{Introduction}
\label{sec:intro}
RGBT tracking has been one of the emerging tasks of the computer vision community, which aims to estimate the position and scale of a pre-labeled object in a video sequence~\cite{rgbt234}. It has diverse applications in robotics, intelligent surveillance, transportation management, and unmanned vehicles ~\cite{lasher}.

\begin{figure}[t]
    \centering

    \includegraphics[scale=0.38]{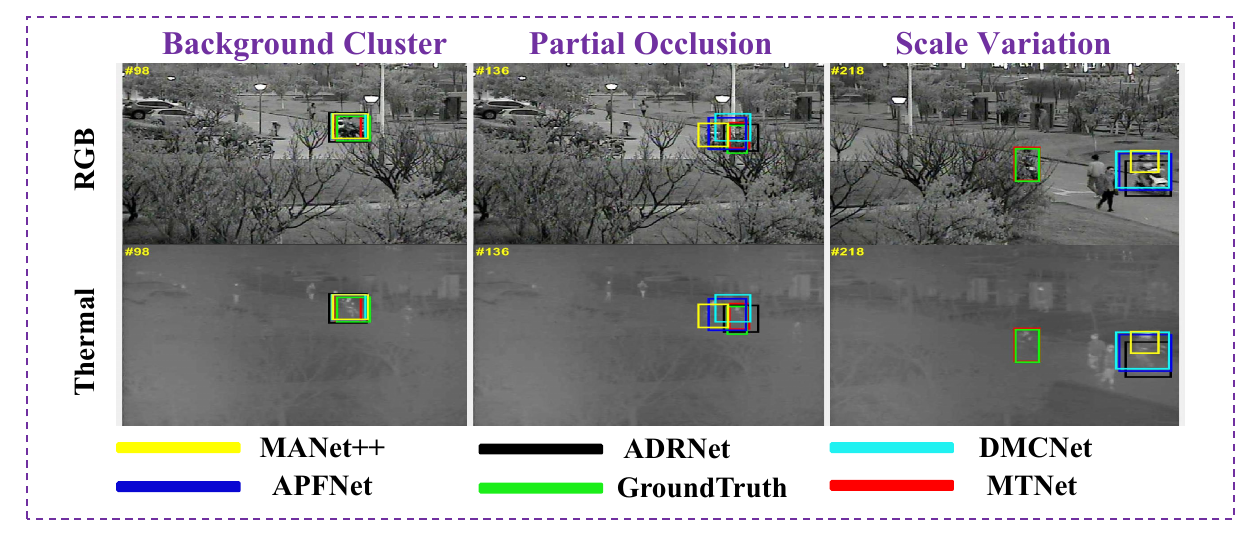}

    \vspace{-0.3cm}
    \caption{Comparison results with representative trackers, \emph{i.e.}, MANet++~\cite{manet++}, ADRNet~\cite{ADRNet}, DMCNet~\cite{dmcnet}, APFNet~\cite{apfnet}. The MTNet performs well in complex scenarios.}
    \label{fig:1}
    \vspace{-0.6cm} 
\end{figure}

Recently trackers based on multi-domain learning~\cite{rtmdnet} seek to enrich target expression by inserting hierarchical feature extraction~\cite{dmcnet,HMFT}, diverse attention mechanisms~\cite{agminet,CMPP} and attribute-aware subnetworks~\cite{apfnet, ADRNet}. Another type is inspired by similarity learning~\cite{siamcda}, which tends to achieve fast speed. Subsequently, the latest transformer-based method~\cite{kbs22} is proposed to push tracking performance to a new level. Nevertheless, the robust feature representation and potential inter-frame information are not explored well due to the regular fusion network and static tracking template. Some visualization examples indicate most trackers still suffer from challenging factors, as shown in Fig.~\ref{fig:1}.

Building on the above analysis, we propose a novel RGBT tracker named MTNet, which addresses two issues as follows: (1) \emph{How to efficiently extract discriminative cues from heterogeneous modalities conducive to instance representation}. (2) \emph{How to estimate the precise bounding box and tackle the tracking challenges}. 

\begin{figure}[t]
    \centering

    \includegraphics[scale=0.6]{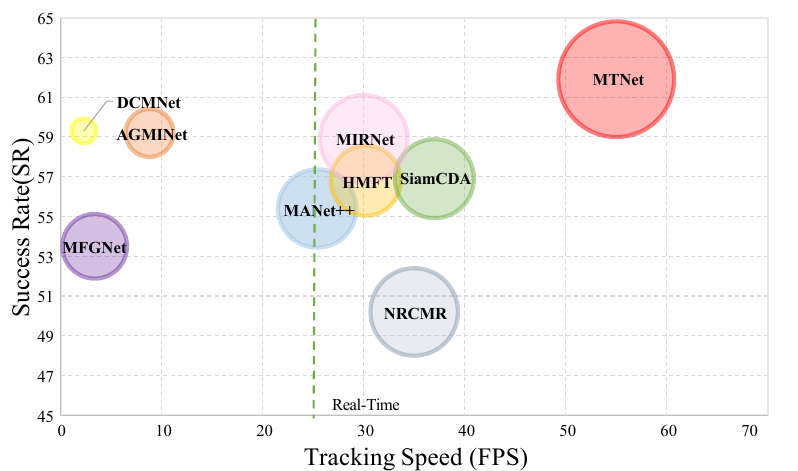}

    \vspace{-0.3cm}
    \caption{Comparison with state-of-the-art trackers, \emph{i.e.}, DMCNet~\cite{dmcnet}, MIRNet~\cite{hou2022mirnet}, MANet++~\cite{manet++}, AGMINet~\cite{agminet}, MFGNet~\cite{mfgnet}, SiamCDA~\cite{siamcda}, NRCMR~\cite{NRCMR}, HMFT~\cite{HMFT} on RGBT234. We plot the Success Rate with respect to the Frames Per Second (FPS) tracking speed. The bubble area represents the weighted sum of the FPS and SR.}
    \label{fig:2}
    \vspace{-0.6cm} 
\end{figure}

For the first issue, we design a modality-aware network to adequately emphasize meaningful features of individual patterns from multiple perspectives. It contains two cost-effective components, \emph{i.e.}, CADM and SSPM, that make the utmost use of attention schemes for robust feature learning. Unlike existing fusion approaches~\cite{siamcda, kbs22}, CADM aims to produce channel-refined features and SSPM flexibly encodes spatial similarity to guide specific modal enhancement. Then, a hybrid attention-based transformer is applied to produce a correlation between the fused template and search region, which comprehensively considers global dependencies via self-attention and cross-attention.
For the second issue, we define a mutual constraint loss by attaching an extra localization branch to establish the associations between the classification and regression branches for joint learning, ensuring accurate results. Instead of adopting optical flow~\cite{dmcnet,agminet} or sub-networks~\cite{hou2022mirnet} to refine the bounding box, the update strategy attempts to maintain a reliable template for boosting inter-frame communication. Experimental results prove MTNet achieves the best results and the top inference speed of 55 FPS on RGBT234, outperforming the newest trackers by a clear margin, as shown in Fig.~\ref{fig:2}. 

The major contributions of this work are summarized as:

$\bullet$ {We propose a novel RGBT tracker that combines the locality and hierarchy of CNN and the global dependency of the transformer to learn modality-aware representations.}

$\bullet$ {We design a trident prediction head by developing the mutual constraint loss function to improve localization accuracy. It further integrates a state-aware template update strategy to boost tracking performance.}

$\bullet$ {Extensive experiments demonstrate that the MTNet achieves satisfactory results while running at real-time speed.}
\vspace{-0.5cm} 
\section{Related work}
\vspace{-0.2cm} 
\subsection{RGBT tracking}
\vspace{-0.1cm} 
Recently, deep learning held the dominating status of RGBT tracking, which mainly consists of two mainstream frameworks. One type of method is based on tracking by detection. For example, 
Lu et al.~\cite{dmcnet} attempted to exploit useful cues across modalities and relieve the disturbance of background clutter by proposing a duality-gated mutual condition network, which yielded competitive results.
Mei et al.~\cite{agminet} presented an asymmetric network to mine heterogeneous features.
Wang et al.~\cite{mfgnet} conducted a novel dynamic convolutional filter to fuse multi-modal features for robust tracking. 
To cope with multiple challenges, some variants~\cite{apfnet,ADRNet} developed attribute-aware sub-networks to generate modality-specific representations.
Moreover, Zhang et al.~\cite{HMFT} contributed an RGBT UAV tracking dataset and then proposed a baseline HMFT by combining a multi-stage fusion.
Another type of work incorporates similarity learning into tracking, which aims to model the optimum matching relationship between the template and search region. For instance,  
Zhang et al.~\cite{siamcda} designed a Siamese-based RGBT tracker with a complementary-aware multi-modal feature fusion.
Feng et al.~\cite{kbs22} used the transformer to strengthen semantic information. 
There is still room for improvement in aspects of cross-modal and inter-frame cues mining.

\vspace{-0.2cm} 
\subsection{Vision Transformer for Tracking}
\vspace{-0.1cm} 
Transformer was first designed for machine translation tasks and has become the dominant structure in natural language processing. The attention mechanism is the key component in Transformer, which learns to establish dependencies between each element in the sequence~\cite{vaswani2017attention}. Given the success of transformers in computer vision, the latest studies have applied this elegant paradigm to visual tracking. For instance, Meinhardt et al.~\cite{trackformer} defined the tracking-by-attention paradigm and designed an end-to-end transformer tracker for multi-object tracking. Chen et al.~\cite{transt} presented a transformer-based feature fusion method to replace the traditional correlation operation for building the matching relation between the template and search region. Yan et al.~\cite{sttrans} proposed a novel tracker with an encoder-decoder transformer by learning spatial-temporal cues to produce satisfactory tracking results. These groundbreaking works will motivate us to bring advanced transformer architectures to RGBT tracking and promote tracking performance.

\vspace{-0.1cm} 

\section{Methodology}
\vspace{-0.2cm} 
\subsection{Network architecture}
\vspace{-0.1cm}

\begin{figure*}[t]
    \centering
    \includegraphics[scale=0.3]{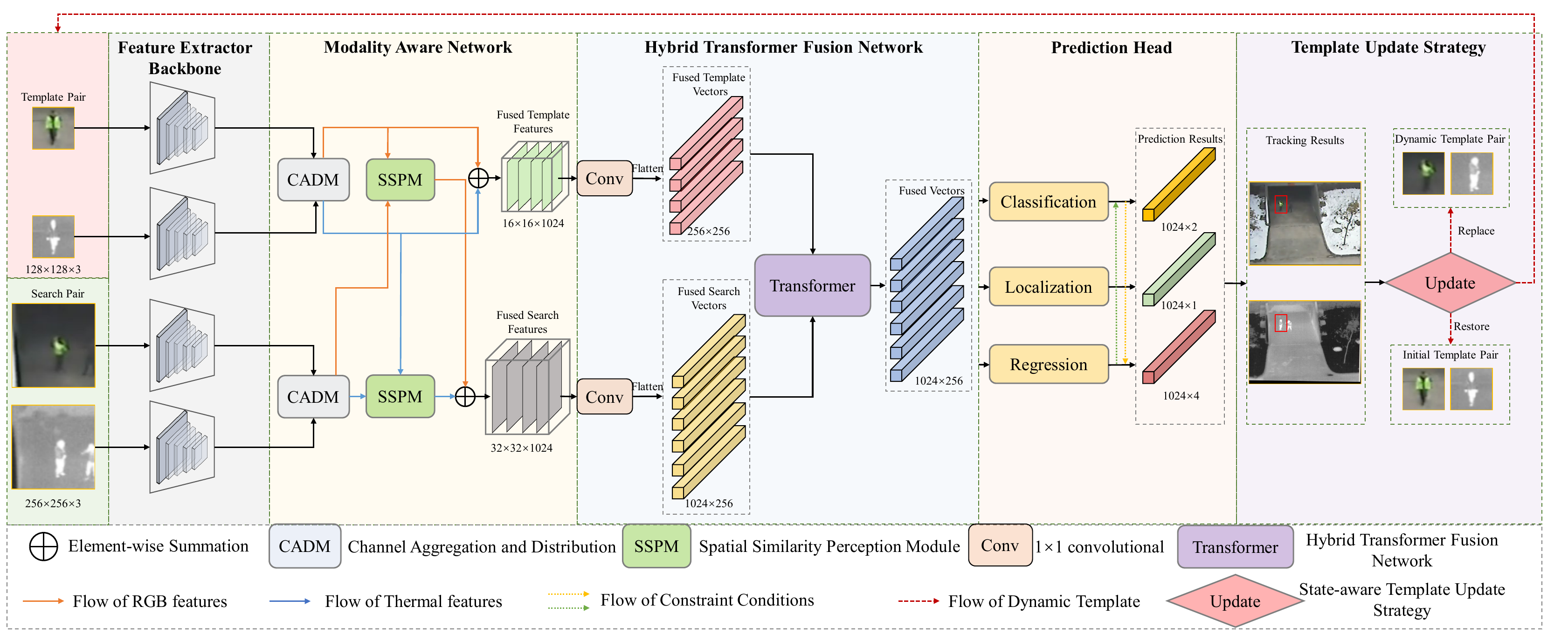}
    \vspace{-0.3cm}
    \caption{Pipeline of MTNet, which is divided into five components, \emph{i.e.}, Feature extractor backbone, Modality-aware network, Hybrid transformer fusion network, Prediction head and Template update strategy.}
    \label{fig:3}
    \vspace{-0.5cm} 
\end{figure*} 

The pipeline is shown in Fig.~\ref{fig:3}. Concretely, we utilize the tailored Resnet-50 as the backbone to obtain the template and search region features.  
Next, the modality-aware network is invented to generate modality-specific representations. To establish the accurate matching correlation, we flatten the fused template and fused search features to vectors and then aggregate them through the hybrid transformer fusion network. Then, the prediction head with triple branches is proposed to estimate the target state. Finally, we apply the template update strategy to select the most appropriate template for refining the subsequent tracking sequences.
\vspace{-0.2cm} 
\subsection{Modality-aware network}
\vspace{-0.1cm} 
\noindent\textbf{Channel aggregation and distribution module.}
\begin{figure}[t]
    \centering
    \includegraphics[scale=0.3]{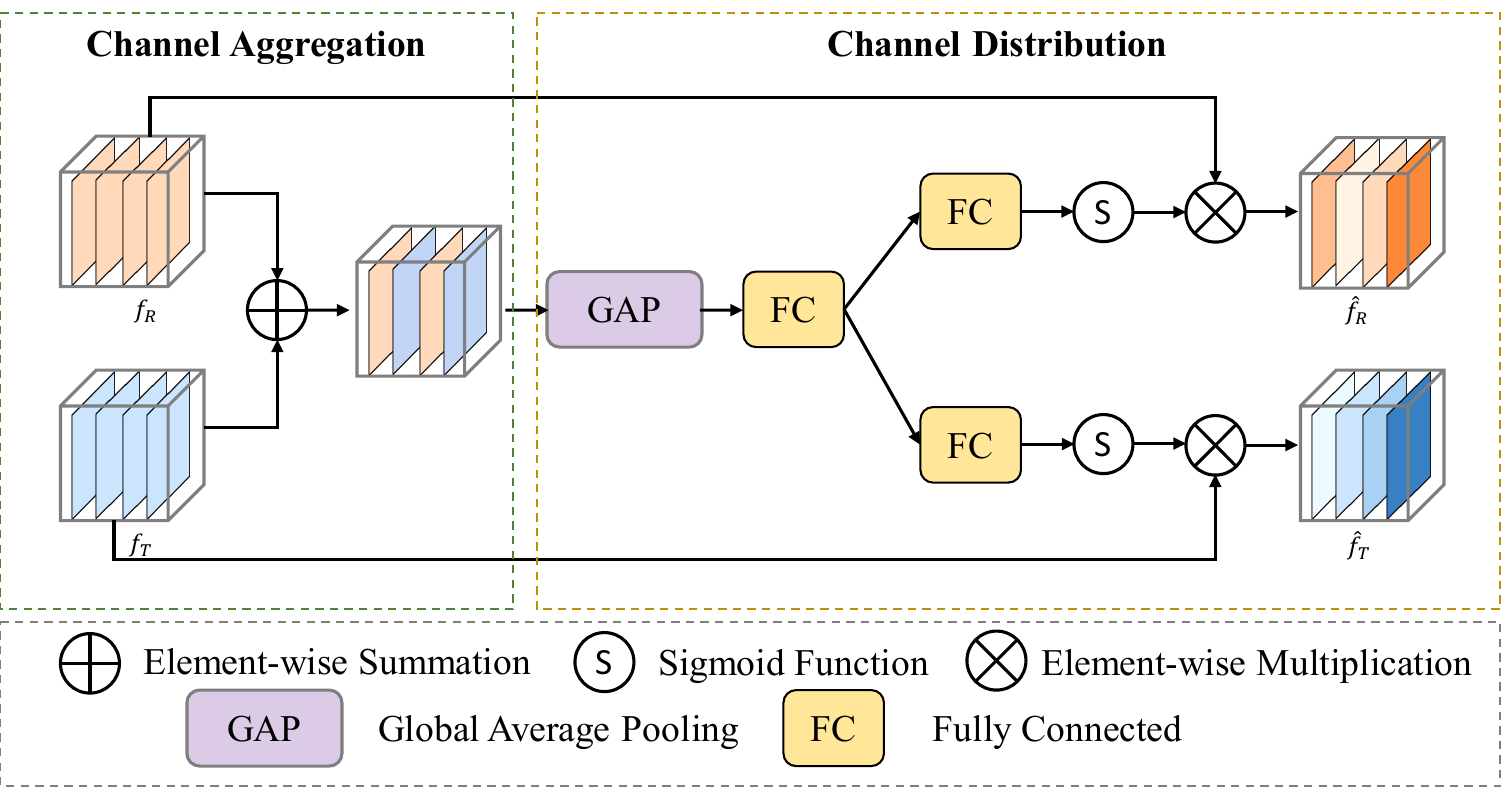}
    \vspace{-0.3cm}
    \caption{Flow chart of the CADM.}
    \label{fig:4}
    \vspace{-0.5cm} 
\end{figure}  
Thermal noise or background clutter is widespread in RGBT data. Channel refinement has become an essential operation in previous works. As shown in Fig.~\ref{fig:4}, we construct a simple yet effective CADM to eliminate the redundant channels of backbone features. In the stage of channel aggregation, we first sum the template features $f_{R}^z$ and $f_{T}^z$ (search region features $f_{R}^x$ and $f_{T}^x$) from the backbone, and then the enriched feature is embedded into the global vector $d_g$ via the Global Average Pooling (GAP) and Fully Connected (FC) layer.
The aggregation operation is defined as
\vspace{-0.2cm} 
\begin{equation}
\ d_g=F_{g}\left(G A P\left(f_{R} \oplus f_{T}\right)\right),
\vspace{-0.2cm} 
\end{equation}
\noindent where $F(\cdot)$ is the FC layer, $R$ and $T$ indicate two modalities of RGB and thermal respectively.
In the stage of channel distribution, we present a two-branch FC layer to get normalization channel-wise weights. Then, the channel-refined template feature maps $\hat{f_{R}^z}$ and $\hat{f_{T}^z}$ are expressed as
\vspace{-0.2cm} 
\begin{equation}
\ \hat{f_{i}^z}={f_{i}^z} \otimes \sigma\left(F_{i}\left(d_g\right)\right), i=R,T, 
\vspace{-0.2cm} 
\end{equation}
where $\otimes$ is element-wise multiplication, $\sigma$ is $Sigmoid$ function. Note that CADM has the same structure but unshared weights for obtaining search region features $\hat{f_{R}^x}$ and $\hat{f_{T}^x}$.

\begin{figure}[t]
    \centering
    \includegraphics[scale=0.3]{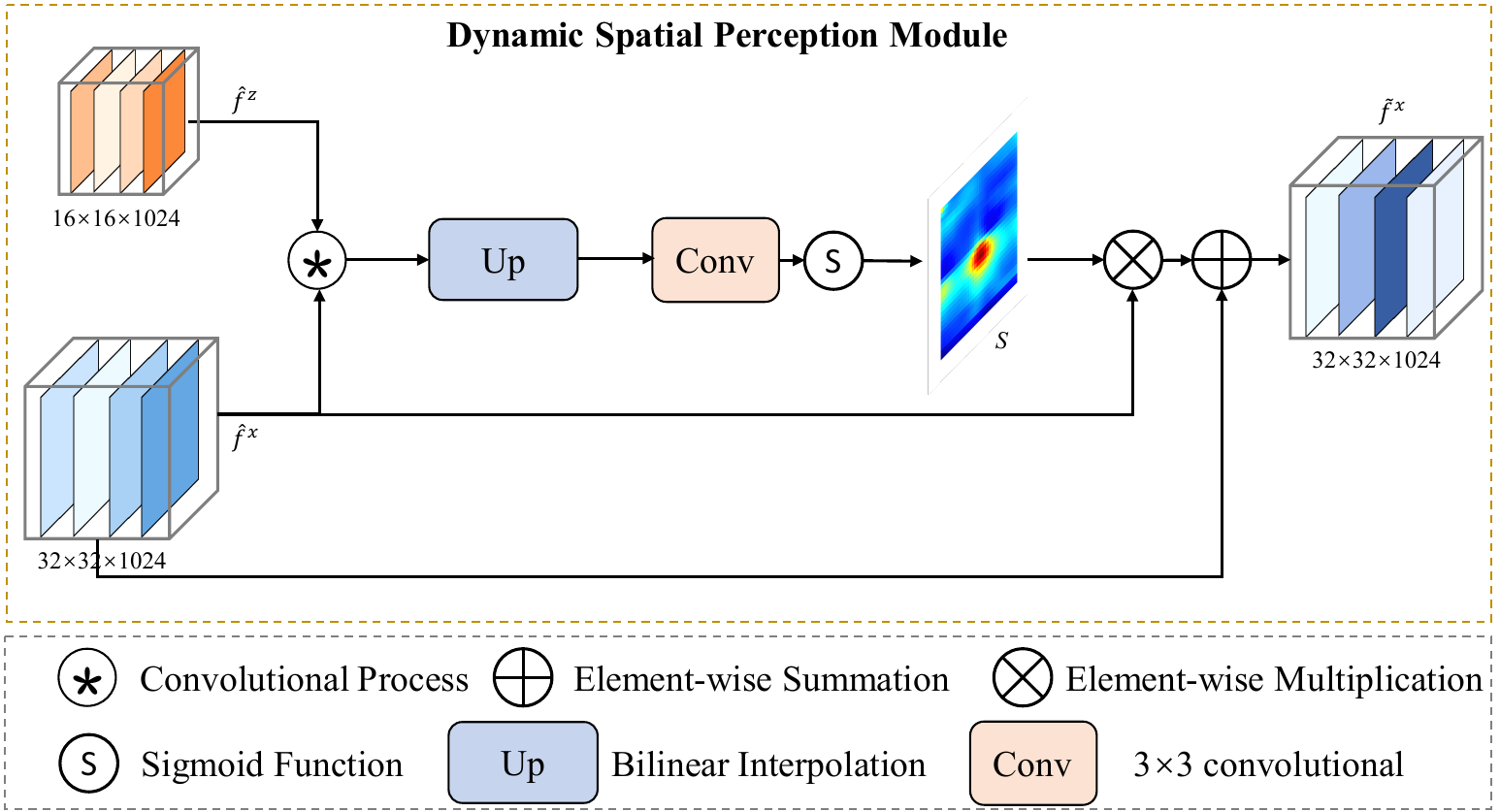}
    \vspace{-0.3cm}
    \caption{Flow chart of the SSPM.}
    \label{fig:5}
    \vspace{-0.5cm} 
\end{figure}

\noindent\textbf{Spatial similarity perception module.} SSPM depends on similarity learning to produce instance-aware residuals for further reinforcing a more reliable pattern. 
The diagram is shown in Fig.~\ref{fig:5}. We first take template features as instance-aware kernels to perform the convolution operation on the corresponding search region and then produce the similarity maps for the two modalities separately. The reason is that the template commonly has a higher responsive intensity in the high-quality search region, and therefore spatial similarity maps are suitable for measuring the reliability of the modality while reinforcing specific representation in the spatial domain. On account of the convolution filtering reducing the resolution of the spatial similarity map, we adopt the bilinear interpolation and convolution operation to refine the spatial similarity map $S_{i}$, which are defined as
\vspace{-0.2cm} 
\begin{equation}
\ S_{i}=\sigma( f_{conv} (up(\hat{f_{i}^z} * \hat{f_{i}^x}))), i=R,T, 
\vspace{-0.2cm} 
\end{equation}
where $*$ denotes convolution operation, $f_{conv}$ means $3 \times 3$ convolution operation, $up$ represents bilinear interpolation, $\sigma$ is the $Sigmoid$ function.
The modality-aware feature maps are generated by attaching the residual connection, and we receive the joint representation by merging each modality feature map. The augmented template and search region features $\tilde{f^z}$ and $\tilde{f^x}$ can be expressed as
\vspace{-0.2cm} 
\begin{equation}
\ \tilde{f^z}=\hat{f_{R}^z} \oplus \hat{f_{T}^z},
\vspace{-0.2cm} 
\end{equation}
\vspace{-0.2cm} 
\begin{equation}
\ \tilde{f^x}=((\hat{f_{R}^x} \otimes S_R) \oplus \hat{f_{R}^x}) \oplus ((\hat{f_{T}^x} \otimes S_T) \oplus \hat{f_{T}^x}).
\vspace{-0.2cm} 
\end{equation}

\vspace{-0.2cm} 
\subsection{Hybrid transformer fusion network}
\begin{figure}[t]
    \centering
    \includegraphics[scale=0.22]{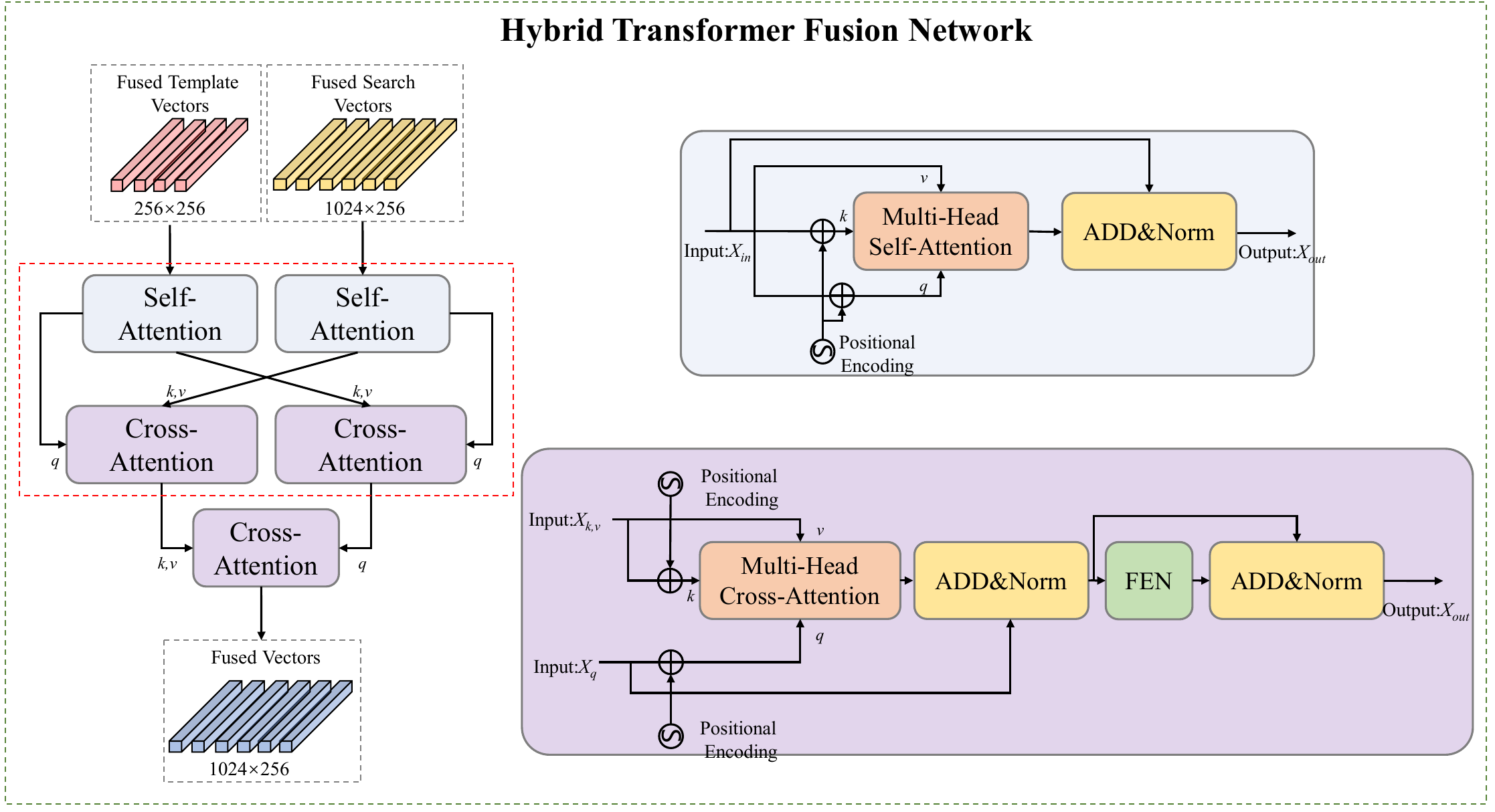}
    \vspace{-0.3cm}
    \caption{Structure of hybrid transformer fusion network.}
    \label{fig:6}
    \vspace{-0.5cm} 
\end{figure}

\vspace{-0.1cm} 
The powerful fusion paradigm from TransT~\cite{transt} is adopted to sense the correlation between the target and search region.
As shown in Fig.~\ref{fig:6}, the template feature $\tilde{f^z}$ and search feature $\tilde{f^x}$ are fed into a $1 \times 1$ convolutional layer and then reshaped to generate two vectors ${X^z}$ and ${X^x}$. Then the transformer fusion network takes ${X^z}$ and ${X^x}$ as the input to mine meaningful features by adopting self-attention and cross-attention. The multi-head cross-attention module aims to fuse features from different branches. Moreover, a feedforward network (FFN) consisting of two linear layers and a Relu activation function boosts the fitting ability of the tracker. We build a hybrid transformer fusion network by stacking those modules four times. Finally, an extra cross-attention module is utilized to obtain the final fused vectors.

\vspace{-0.2cm} 
\subsection{Trident prediction head}
\vspace{-0.1cm} 
The prediction head contains three branches, \emph{i.e.}, classification, regression and localization. 
Due to the prediction inconsistency between classification and regression, we insert a mutual constraint flow into binary cross-entropy loss by multiplying the normalized IoU, which aims to suppress unreasonable proposals. The classification loss is formulated as
\begin{equation}
 L_{cls}=-\sum_j((y_j\log(p_j)IoU+(1-y_j)\log(1-p_j))),
\vspace{-0.2cm} 
\end{equation}
where $y_j$ defines the label of the jth sample $y_j=1$ denotes the positive sample, $p_j$ indicates the probability belonging to the foreground, and IoU represents the Intersection over Union between prediction and ground truth. The regression loss contains two parts: $l_1$-norm loss and Complete IoU loss~\cite{ciou}, which is defined as
\begin{equation}
L_{reg}=\sum_j\Pi_{y_j=1}(\lambda_1L_1(b_j,\hat{b_j})+\lambda_C L_{CIoU}(b_j,\hat{b_j})p_{j}),
\vspace{-0.2cm} 
\end{equation}
where $b_j$ means the j-th bounding box, and $p_j$ denotes the corresponding classification confidence of the positive samples. The regularization parameters $\lambda_1$ and $\lambda_{C}$ are set to 5 and 2 respectively.
Localization loss is constructed by binary cross-entropy loss and is described as
\begin{equation}
L_{loc}=-\sum_j(O_j\log(p^{loc}_j)+(1-O_j)\log(1-p^{loc}_j)),
\vspace{-0.2cm} 
\end{equation}
where $O_j$ denotes the IoU scores calculated by the regression branch, and $p^{loc}_j$ means the predicted value of the localization branch.
The overall loss is calculated as follows
\begin{equation}
Loss=n_1L_{cls}+n_2L_{reg}+n_3L_{loc}.
\vspace{-0.2cm} 
\end{equation}
where $n_1$, $n_2$ and $n_3$ represent the hyperparameters.
\vspace{-0.2cm} 
\subsection{State-aware template update strategy}
\vspace{-0.1cm} 
In practical tracking tasks, the appearance of the target object often changes over time. If the tracking template is not updated in a timely manner, tracking failures can occur. Given the real-time requirements, it is preferable to design a low-cost update strategy instead of relying on an additional auxiliary model. To achieve this, the proposed strategy divides the tracking process into three states based on confidence levels, \emph{i.e.}, steady state, transient steady state, and unstable state. Note that confidence is calculated by multiplying classification scores and localization scores. 
Specifically, the steady state is defined as the condition in which the confidence score of $M$ consecutive frames is greater than 0.9. Once the steady state is reached, the current template will replace the initial template. If the confidence score is between 0.7 and 0.9, we reckon the tracker is in a state of transient steady and the template remains constant during this interval. 
If the confidence is lower than 0.7 and has accumulated up to $N$ times, the tracker may struggle in an unstable state, and the current template is restored by an initial template.
To pursue the best performance, we set different update intervals for each state.

\begin{table*}[t]
\footnotesize
\begin{center}
\caption{Comparison results of our method against the state-of-the-art trackers. Attribute-based and overall performance are evaluated by PR/SR scores(\%) and are produced on RGBT234. The best and second best results are in \textcolor{red}{red} and \textcolor{green}{green}.} \label{tab:1}
\begin{tabular}{cccccccccc}
\hline
Trackers             & DMCNet~\cite{dmcnet}   & MIRNet~\cite{hou2022mirnet}    & APFNet~\cite{apfnet}    & AGMINet~\cite{agminet}       & MFGNet~\cite{mfgnet}     & SiamCDA~\cite{siamcda}    & RMWT~\cite{kbs22}    & HMFT~\cite{HMFT}     & \textbf{MTNet} \\ \hline
Pub. Info.          & TNNLS2022   & ICME2022    & AAAI2022    & TIM2022       & TMM2022     & TCSVT2022    & KBS2022    & CVPR2022     & -\\ \hline
NO           & 92.3 / 67.1 & \textcolor{red}{95.4} / \textcolor{red}{72.4} & 94.8 / 68.0 & \textcolor{green}{94.9} / 69.1 & 92.0 / 64.0 & 88.4 / 66.4 & 92.1 / \textcolor{green}{70.8} & 90.9 / 67.4 & 91.0 / 67.8               \\
PO           & \textcolor{green}{89.5} / 63.1 & 86.1 / 62.7 & 86.3 / 60.6 & \textcolor{red}{90.2} / 63.9 & 84.3 / 58.0 & 84.2 / \textcolor{green}{63.9} & 85.4 / 63.6 & 85.7 / 62.1 & 88.7 / \textcolor{red}{64.8}               \\
HO           & 74.5 / 52.1 &  71.0 / 49.0 & 73.8 / 50.7 & 72.9 / 50.3 & 66.2 / 44.3 & 66.2 / 48.7 & \textcolor{green}{75.2} / \textcolor{green}{55.5} & 66.4 / 46.9 & \textcolor{red}{78.6} / \textcolor{red}{56.3}                \\
LI           & \textcolor{green}{85.3} / 58.7 & 83.4 / 57.5 & 84.3 / 56.9 & \textcolor{red}{87.0} / \textcolor{green}{59.8} & 79.1 / 54.2 & 81.8 / 61.2 & 84:1 / \textcolor{red}{61.5} & 83.3 / 59.1 & 83.3 / 59.5                \\
LR           & \textcolor{green}{85.4} / \textcolor{red}{57.9} & 83.9 / 56.3 & 84.4 / 56.5 & \textcolor{red}{86.7} / \textcolor{green}{57.2} & 79.3 / 49.5 & 70.9 / 49.9 & 76.6 / 55.0 & 76.3 / 57.1 & 80.4 / 55.4                \\
TC           & \textcolor{red}{87.2} / \textcolor{green}{61.2} & 81.1 / 59.1 & 82.2 / 58.1 & 80.6 / 59.2 & 81.8 / 55.8 & 67.4 / 47.7 & 78.2 / 58.6 & 72.2 / 50.4 & \textcolor{green}{86.1} / \textcolor{red}{61.6}           \\
DEF          & 77.9 / 56.5 & 77.8 / 58.1 & 78.5 / 56.4 & 79.5 / 56.8 & 72.1 / 50.8 & 77.9 / 59.2 & \textcolor{green}{80.3} / \textcolor{green}{62.0} & 77.6 / 57.9 & \textcolor{red}{84.7} / \textcolor{red}{64.0}            \\
FM           & \textcolor{red}{80.0} / 52.4 & 68.3 / 47.1 & 79.1 / 51.1 & \textcolor{green}{79.4} / 51.2 & 72.5 / 44.6 & 61.4 / 45.3 & 74.3 / \textcolor{green}{55.3} & 65.9 / 46.9 & 79.2 / \textcolor{red}{58.0}           \\
SV           & 84.6 / 59.8 & 82.7 / 61.9 & 83.1 / 57.9 & 83.2 / 59.3 & 76.1 / 52.8 & 77.7 / 59.3 & \textcolor{green}{86.1} / \textcolor{green}{65.9} & 80.0 / 59.2 & \textcolor{red}{89.0} / \textcolor{red}{66.1}            \\
MB           & 77.3 / 55.9 & 74.6 / 54.6 & 74.5 / 54.5 & \textcolor{green}{78.2} / 57.5 & 73.7 / 51.0 & 63.6 / 47.9 & 76.8 / \textcolor{green}{57.8} & 70.6 / 50.9 & \textcolor{red}{83.4} / \textcolor{red}{61.6}           \\
CM           & 80.1 / 57.6 & 76.4 / 55.4 & 77.9 / 56.3 & 79.0 / 57.5 & 73.2 / 50.4 & 73.3 / 54.7 & \textcolor{green}{83.1} / \textcolor{green}{62.7} & 77.9 / 56.2 & \textcolor{red}{86.0} / \textcolor{red}{63.4}            \\
BC           & \textcolor{red}{83.8} / \textcolor{red}{55.9} & 78.9 / 51.7 & 81.3 / 54.5 & \textcolor{green}{83.3} / \textcolor{green}{55.3} & 74.3 / 45.9 & 74.0 / 52.9 & 74.5 / 52.5 & 73.8 / 49.8 & 74.9 / 50.8            \\
\textbf{ALL} & 83.9 / 59.3 & 81.6 / 58.9 & 82.7 / 57.9 & \textcolor{green}{84.0} / 59.2 & 78.3 / 53.5 & 76.0 / 56.9 & 82.5 / \textcolor{green}{61.6} & 78.8 / 56.8 & \textcolor{red}{85.0} / \textcolor{red}{61.9}            \\ \hline

\end{tabular}
\end{center}
\vspace{-0.5cm} 
\end{table*}

\vspace{-0.2cm}
\section{Experiments}
\vspace{-0.1cm} 
\subsection{Datasets and metrics}
\vspace{-0.1cm} 
In this paper, we conduct comparative experiments with high-performance competitors on three popular RGBT benchmarks, \emph{i.e.}, GTOT~\cite{gtot}, RGBT234~\cite{rgbt234} and LasHeR~\cite{lasher}. Following mainstream works, we employ two classical metrics, Precision Rate (PR) and Success Rate (SR) to measure tracking performance. For a fair comparison, we set the threshold of GTOT to 5 pixels and RGBT234/LasHeR to 20 pixels considering the inherent inconsistent image resolution between different datasets. Moreover, we apply the Normalized Precision Rate (NPR) metric to alleviate the influence of the resolution for testing the LasHeR.
\vspace{-0.2cm} 
\subsection{Implementation Details}
\vspace{-0.1cm} 
The MTNet is implemented on the PyTorch 1.10 platform with an E5-2680V4 CPU, 64GB RAM, and two NVIDIA GeForce RTX3090 GPUs with 24GB memory. In the offline training phase, MTNet is trained on the LasHeR. The AdamW optimizer is utilized to update the model, and the initial learning rate and weight decay are both set to $1e^{-4}$. 
The model converges within 40 epochs, and each epoch contains 1000 iterations. The batch size is set to 16. After 20 epochs, the learning rate is decreased by a factor of 10. We set the hyperparameters $n_1$, $n_2$ and $n_3$ to 8, 5 and 1 respectively. In the online tracking phase, the prediction head outputs 1024 proposals, which are ranked with the window penalty and location logits. We set the update intervals $\{M, N\}$ as $\{50, 2\} $for testing the GTOT. For evaluating other datasets, the parameters are set as $\{70, 2\}$.
Finally, the bounding box with the maximum confidence is regarded as the best tracking result.

\begin{table}[htbp]
\vspace{-0.5cm}
\begin{center}
\footnotesize
\caption{Comparison results on GTOT.} \label{tab:2}
\begin{tabular}{cccccc}
\hline
Trackers      & HMFT~\cite{HMFT} & DMCNet~\cite{dmcnet} &  CMPP~\cite{CMPP}   &  MTNet  \\ \hline
PR &  91.3  & 90.9  & 92.6     & {93.5}\\
SR &  74.9  & 73.3  & 73.8     & {76.0}\\
 \hline
\end{tabular}
\end{center}
\vspace{-0.8cm} 
\end{table}

\vspace{-0.3cm} 
\subsection{Comparison with the state-of-the-art}
\vspace{-0.2cm} 
The proposed tracker is compared with 11 latest methods, \emph{i.e.}, DMCNet~\cite{dmcnet}, MIRNet~\cite{hou2022mirnet}, APFNet~\cite{apfnet}, AGMINet~\cite{agminet}, MFGNet~\cite{mfgnet}, SiamCDA~\cite{siamcda}, RMWT~\cite{kbs22}, HMFT~\cite{HMFT}, CMPP~\cite{CMPP}, MANet++~\cite{manet++}, ADRNet~\cite{ADRNet}.

\begin{figure}[t]
\vspace{-0.3cm} 
    \centering
    \includegraphics[scale=0.25]{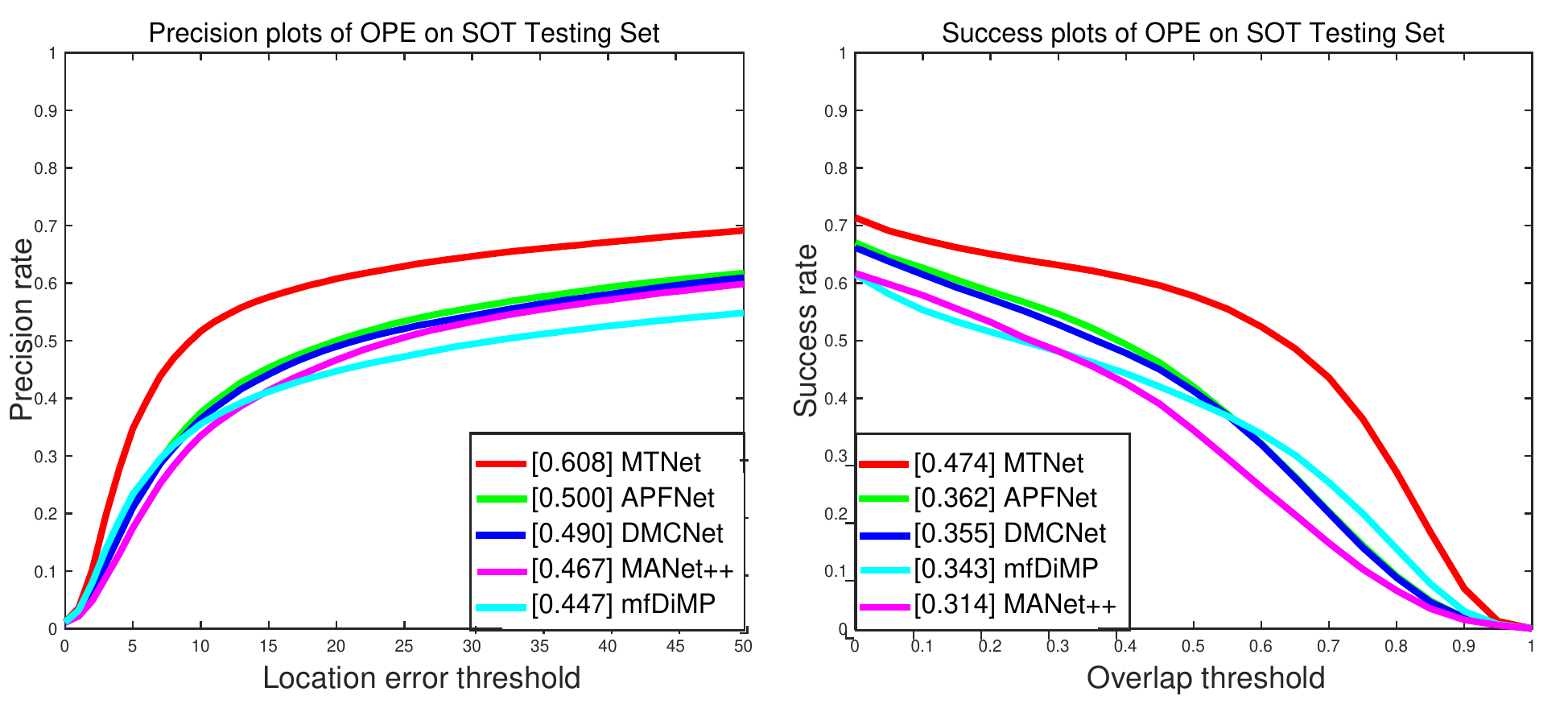}
    \vspace{-0.3cm}
    \caption{The evaluation curve on LasHeR.}
    \label{fig:7}
    \vspace{-0.3cm} 
\end{figure}

\noindent\textbf{Overall performance.} As reported in Table~\ref{tab:1}, MTNet outperforms all other competitors with 85.0\%/61.9\% in PR/SR, achieving performance gains of 1.0\%/2.7\% over the second-ranked tracker AGMINet on RGBT234. Besides, MTNet achieves the best results on GTOT, with PR/SR reaching 93.5\%/76\%, as given in Table~\ref{tab:2}. Specifically, MTNet outperforms the HMFT by 2.2\%/1.1\% in PR/SR. 
Fig.~\ref{fig:7} shows that MTNet obtains the best ranking on LasHeR, with 60.9\%, 56.3\% and 47.4\% in PR, NPR and SR. Furthermore, even when compared to the retrained tracker mfDiMP~\cite{DIMP} for comparison, MTNet still surpasses it by 0.9\%/0.7\%.

\noindent\textbf{Attribute-based performance.} The attribute-based comparison on RGBT234 are presented in Table~\ref{tab:1}. The attributes consist of no occlusion(NO), partial occlusion(PO), heavy occlusion(HO), low illumination(LI), deformation(DEF), fast motion(FM), scale variation(SV), motion blur(MB), camera moving(CM), low resolution(LR), thermal crossover(TC) and background cluster(BC). Experimental results suggest that the proposed approach works well in adverse conditions. 
Compared to the transformer-based method RMWT, MTNet achieves performance gains of 3.4\%/0.8\%, 4.7\%/2\%, 2.9\%/0.2\% and 6.6\%/3.8\% in the attributes of HO, DEF, SV and MB, respectively. One of the important reasons for the superior performance of MTNet is the incorporation of CADM and SSPM modules, which enable the learning of robust multi-modal representations. Additionally, the proposed state-aware template update strategy helps to mitigate the impact of unreliable appearance features. In the attributes of NO, LI, LR, and BC, AGMINet, and DMCNet achieve the best performance due to their online training and extra refinement network, but at the cost of increased complexity.
\vspace{-0.1cm} 
\vspace{-0.2cm} 
\subsection{Ablation study}
\vspace{-0.1cm} 
\noindent\textbf{Components analysis.} To further validate the feasibility of each contribution, we implement three variants and test them on LasHeR datasets, \emph{i.e.}, (I) is a base model, which is by TransT~\cite{transt} and integrates dual-pattern features via simple element addition. (II) incorporates the modality-aware network into the baseline. (III) combines prediction head with mutual constraint loss on the basis of II. (IV) is the final version equipped with the template update strategy.
According to the tracking results reported in Table~\ref{tab:2}, we can draw the following conclusions: 1) The modality-aware network flexibly exploits cues between dual patterns to enhance modality-aware representation. 2) The trident prediction head improves localization accuracy by unifying the distribution between each branch. 3) The template update strategy introduces additional temporal context to alleviate the appearance variation issue.

\begin{table}[t]
\vspace{-0.3cm}
\begin{center}
\footnotesize
\caption{Ablation study on different components.} \label{tab:3}
\begin{tabular}{ccccccc}
\hline
Variants      & Modality-aware & Loss &  Update & PR &    NPR   &  SR \\ \hline
I &    &   &  & 56.8  & 52.4  & 44.9 \\
II & \checkmark   &   & & 58.6  & 54.1  & 46.2 \\
III & \checkmark   & \checkmark  &  & 59.4  & 55.0  & 46.5 \\
IV & \checkmark   & \checkmark  & \checkmark & 60.8  & 56.3  & 47.4 \\ \hline
\end{tabular}
\end{center}
\vspace{-0.7cm} 
\end{table}
\begin{table}[ht]
\vspace{-0.2cm}
\begin{center}
\footnotesize
\caption{Comparison of different thresholds on RGBT234.} \label{tab:4}
\begin{tabular}{cccc}
\hline
Update interval      & $N=0$ & $N=2$ & $N=5$     \\ \hline
$M=60$    & 83.3 / 60.5    & 83.7 / 60.8  & 84.2 / 60.5 \\ 
$M=70$    & 84.7 / 61.7    & \textbf{85.0} / \textbf{61.9}  & 84.9 / 61.8 \\ 
$M=80$    & 84.0 / 61.1    & 83.7 / 60.9  & 84.0 / 61.1 \\ \hline
\end{tabular}
\end{center}
\vspace{-0.8cm} 
\end{table}
\begin{figure*}[t]
    \centering
    \includegraphics[scale=0.33]{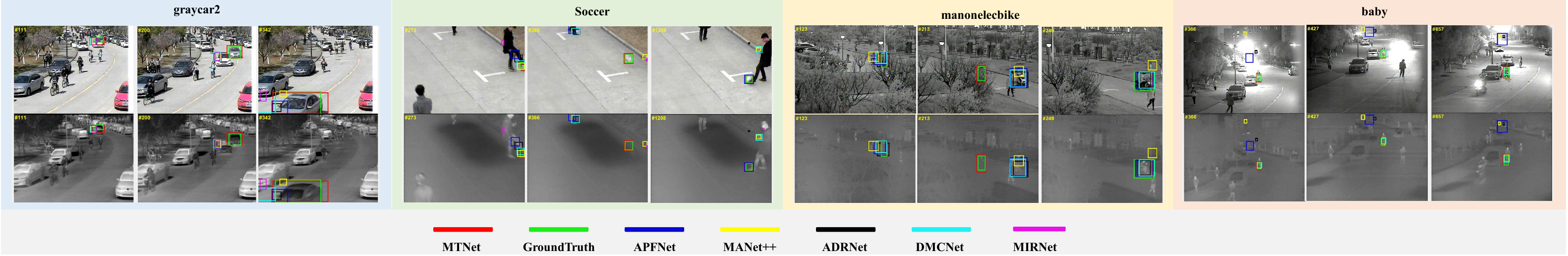}
    \vspace{-0.3cm}
    \caption{Qualitative comparison between MTNet and other trackers on four challenging sequences.}
    \label{fig:8}
    \vspace{-0.5cm} 
\end{figure*}  

\noindent\textbf{Parameters analysis.} To measure the impact on performance, we set $M = \{60, 70, 80\}$, $N = \{0, 2, 5\}$ to carry out evaluation and the comparison results on RGBT234 are reported in Table~\ref{tab:4}. We observe the best metrics are determined by the parameter $\{70,2\}$. When the interval is longer or shorter it may lead to a suboptimal template. In addition, instantaneously resetting the current template may cause misjudgment of the state. In both cases, the best results may not be achieved. Hence, selecting an appropriate update interval can effectively improve tracking performance. Since the scale of GTOT is small, we set the parameters as $\{50,2\}$.

\noindent\textbf{Qualitative Analysis}.
The qualitative comparison is shown in Fig.~\ref{fig:8}. Thanks to the modality-aware representation and reliable template, the proposed tracker performs well when encountering multiple challenges, especially FM, HO, SV, and MB. Therefore, the superiority of MTNet has been adequately verified again via intuitive qualitative comparison.
\vspace{-0.5cm} 
\section{Conclusion}
\vspace{-0.3cm} 
In this work, we proposed a novel MTNet for robust RGBT tracking. A modality-aware network was invented to reinforce modality-specific cues from multiple perspectives, while a hybrid transformer fusion network was utilized to establish the long-distance association between the augmented features. The trident prediction head and the state-aware template update strategy were jointly used to a high-quality dynamic template that tackles various tracking challenges and realizes stable all-weather tracking. Experiments verify that the proposed method attains impressive performance compared to state-of-the-art trackers while achieving top speed.

\vspace{-0.5cm}

% References should be produced using the bibtex program from suitable
% BiBTeX files (here: strings, refs, manuals). The IEEEbib.bst bibliography
% style file from IEEE produces unsorted bibliography list.
% -------------------------------------------------------------------------
\bibliographystyle{IEEEbib}
\bibliography{icme2023template}

\end{document}